\documentclass[dvipsnames,format=sigconf,anonymous=false,review=false,nonacm]{acmart}

\usepackage{subfig}
\usepackage[noend]{algorithmic}
\usepackage{algorithm}
\usepackage{lipsum} 

\AtBeginDocument{%
  \providecommand\BibTeX{{%
    \normalfont B\kern-0.5em{\scshape i\kern-0.25em b}\kern-0.8em\TeX}}}

\setcopyright{none}

\begin{document}

\title{Minimum variance threshold for \texorpdfstring{$\epsilon$}{epsilon}-lexicase selection}

\author{Guilherme Seidyo Imai Aldeia}
\orcid{0002-0102-4958}
\affiliation{%
  \institution{Federal University of ABC}
  \city{Santo Andre}
  \state{São Paulo}
  \country{Brazil}
}
\email{guilherme.aldeia@ufabc.edu.br}

\author{Fabrício Olivetti de França}
\orcid{0000-0002-2741-8736}
\affiliation{%
  \institution{Federal University of ABC}
  \city{Santo Andre}
  \state{São Paulo}
  \country{Brazil}
}
\email{folivetti@ufabc.edu.br}

\author{William G. La Cava}
\orcid{0000-0002-1332-2960}
\affiliation{%
  \institution{Boston Children's Hospital}
  \institution{Harvard Medical School}
  \city{Boston}
  \state{Massachusetts}
  \country{USA}
}
\email{william.lacava@childrens.harvard.edu}

\begin{abstract}
Parent selection plays an important role in evolutionary algorithms, and many strategies exist to select the parent pool before breeding the next generation.
Methods often rely on average error over the entire dataset as a criterion to select the parents, which can lead to an information loss due to aggregation of all test cases.
Under $\epsilon$-lexicase selection, the population goes to a selection pool that is iteratively reduced by using each test individually, discarding individuals with an error higher than the elite error plus the median absolute deviation (MAD) of errors for that particular test case.
In an attempt to better capture differences in performance of individuals on cases, we propose a new criteria that splits errors into two partitions that minimize the total variance within partitions.
Our method was embedded into the FEAT symbolic regression algorithm, and evaluated with the SRBench framework, containing $122$ black-box synthetic and real-world regression problems.
The empirical results show a better performance of our approach compared to traditional $\epsilon$-lexicase selection in the real-world datasets while showing equivalent performance on the synthetic dataset.
\end{abstract}



\maketitle

\section{Introduction}

Genetic Programming (GP) \cite{koza1994genetic} is a meta-heuristic created to evolve computer programs to solve a particular problem, inspired by a weak metaphor of the biological evolutionary process.
One application that is particularly suited for GP is symbolic regression (SR)~\cite{koza1992genetic,srbench}, a supervised learning method that searches for a function in the parametric family of functions and the optimal parameter that best fits a set of observed examples. Given a set of $d$ inputs of dimension $\mathbf{x} \in \mathbb{R}^n$ and respective target values $\{\mathbf{x}_i, y_i\}_{i=1}^d = (\mathcal{X}, \mathbf{y})$, SR optimizes both the structure and parameters of a model, $\hat{f}(\mathbf{X},\hat{\theta})$.

An essential aspect of any evolutionary algorithm is the selection of solutions that will be recombined to generate new \emph{offspring} solutions. The main goal of the recombination operator is to combine different parts of highly fit parents with the expectation that the offspring will improve upon its parents. As such, this selection requires that the parents perform well at different dataset examples.
Traditional approaches such as tournament selection \cite{miller1995genetic, brindle1980genetic} only use the aggregated information the fitness function provides to decide which parents will reproduce. As such, it can choose parents that perform well on average across the whole dataset while missing individuals that perform well on difficult subsets of the problem.
Unsurprisingly, evidence shows that parent selection with the entire data ignores relevant cases, as less information is available \cite{10.1145/2576768.2598288}.

An alternative approach is to make the selection by distinguishing between the individuals by rewarding those who excel in difficult subsets of test cases. This is the idea behind the lexicase selection \cite{10.1145/2330784.2330846}, which has shown state-of-the-art performance in program synthesis and symbolic regression \cite{ding_probabilistic_2023}.
It is based on the assumption that problem modularity is identifiable to some degree by its individual fitness cases, each representing a circumstance in which the program must do well.
To select one parent, the entire population goes into a selection pool. Then, a random test case is selected, and individuals who fail that test are eliminated from the pool. Randomly selecting the following test case is repeated until the pool contains only one individual. If the procedure runs out of test cases, a random individual from the pool is returned with uniform probability.

While the original lexicase is suitable for binary cases (correct or incorrect test cases), it required modifications to handle continuous values in the SR context by defining a threshold $\epsilon$ \cite{la_cava_epsilon-lexicase_2016}. This has shown to be effective for regression compared to other selection techniques. For each test case, a threshold is established based on the median absolute deviation of errors, and any individual within the tolerance remains in the pool.

This paper proposes a new way of estimating $\epsilon$ by replacing the median absolute deviation with a threshold that splits the errors into two partitions, such that the total sum of variance for the partitions is minimized.
This proposed threshold criterion corresponds to a type of information entropy, used commonly by decision tree methods \cite{breiman1984classification}. By minimizing the number of different individuals on either side, it behaves as a $1$-dimensional clustering.
This represents a way of picking the individuals that remain in the pool that is more sensitive to similarities in performance among the pool.
 
We evaluate the proposed method by implementing it into FEAT \cite{la_cava_learning_2019}, a symbolic regression framework that has been previously evaluated with $\epsilon$-lexicase selection \cite{srbench} and successfully applied to real-world problem \cite{la_cava_flexible_2023}. Our experiments are twofold: first, with $6$ low-dimensional datasets, we provide an in-depth analysis of its performance characteristics during the evolution. Subsequently, we rigorously assessed the proposed selection process utilizing the SRBench \cite{srbench}, a unified framework designed to perform a relative performance evaluation of symbolic regression algorithms.

Our empirical findings show that our method presents a superior performance in terms of $R^2$ while keeping solutions within the same level of complexity when compared to the $\epsilon$-lexicase selection. 
In SRBench, our method improved the FEAT algorithm in real-world problems and was still capable of increasing its ranking.

The remainder of the paper is organized as follows.
\S\ref{sec:split-threshold} details our proposed modification by replacing the median absolute deviation with a more effective threshold optimization strategy.
\S\ref{sec:feat} presents the symbolic regression framework used in the experiments.
\S\ref{sec:methods} outlines experimental design, datasets, and evaluation criteria.
Results and discussion are presented in \S\ref{sec:results-discussion} offering a comparative analysis of our approach against $\epsilon$-lexicase. Finally, \S\ref{sec:conclusions} summarizes findings, discusses implications, and suggests future avenues for research in evolutionary algorithms and parent selection methodologies.

\section{\texorpdfstring{$\epsilon$}{epsilon}-lexicase selection}~\label{sec:epsilon-lexicase}

Let us define a test case as a pair $(\mathbf{x}_i, y_i)$, $i=1, 2, \ldots, d$ from the data split used during fitness evaluation. We denote $t \in \mathcal{T}$ any possible test case. The error of an individual $n$, denoted as $e_t(n)$, is the absolute difference between its predicted value given $\mathbf{x}_t$ and the target value $y_t$. The vector of errors for a test case $t$, denoted as $\mathbf{e}_t$, is the concatenation of errors for all individuals from the set. The set $\mathcal{N}$ denotes the population, and the set $\mathcal{S}$ denotes the selection pool.
Initially, La Cava et al. \cite{la_cava_epsilon-lexicase_2016} proposed using a threshold in lexicase to handle continuous values. The method was further studied by them in later work \cite{la_cava_probabilistic_2018}, from which different implementations of $\epsilon$-lexicase were proposed: static, semi-dynamic, and dynamic.

In the static scenario, the criteria to stay in the pool uses $e_t^* + \epsilon$, where $e_t^*$ is the best error from the population to that case.
In semi-dynamic and dynamic implementations, the threshold to stay in the pool is calculated as the error of the $\mathtt{elite}$ (smallest error from the pool) plus $\epsilon$. The difference between the semi-dynamic and the dynamic is that the former calculates $\epsilon$ over the entire population, while the latter calculates it for the pool.
In any case, $\epsilon$ is calculated as the median absolute deviation (MAD) \cite{PHAMGIA2001921}:

\begin{equation}
    \epsilon_t = \lambda(\mathbf{e}_t) = \text{median}(|\mathbf{e}_t - \text{median}(\mathbf{e}_t)|).
\end{equation}

The MAD produces median-centered random variables with more robustness than standard deviation, as the median is more representative of the center than the mean in asymmetrical distributions \cite{PHAMGIA2001921}. The interpretation of the MAD as a criterion in the $\epsilon$-lexicase context is a dispersion measured from the center of the distribution.

\begin{table*}[tbh]
\centering
\caption{Comparison of $\epsilon$-lexicase variations.}
\label{tab:comparison_e_lexicases}
\begin{tabular}{@{}llll@{}}
\toprule
\textbf{Strategy} &
  \textbf{Pool} &
  \textbf{Criteria} &
  \textbf{Characteristic} \\ \midrule
Static &
  pop ($\mathcal{N}$) &
  $e_t^* + \epsilon_t$ &
  \begin{tabular}[c]{@{}l@{}} The error vector is calculated previously with entire population. A binary mask\\ matrix stores the global criteria to iterate through the random cases\end{tabular} \\ \hline
Semi-dynamic &
  pop ($\mathcal{N}$) &
  \begin{tabular}[c]{@{}l@{}}$\texttt{elite} + \epsilon_t$,\\ where $\mathbf{\epsilon} \leftarrow \lambda(\textbf{e}_t) \text{ for } t \in \mathcal{T}$\end{tabular} &
  \begin{tabular}[c]{@{}l@{}} The error vector is calculated previously, but the decision criteria is based on the\\ individuals that are in the pool, as it uses the elite error in the criteria\end{tabular} \\ \hline
Dynamic &
  pool ($\mathcal{S}$) &
  \begin{tabular}[c]{@{}l@{}}$\texttt{elite} + \epsilon_t$,\\ where $\mathbf{\epsilon} \leftarrow \lambda(\textbf{e}_t(\mathcal{S}))$\end{tabular} &
  Criteria is dependent of the elite, and $\lambda$ is calculated with the selection pool \\ \bottomrule
\end{tabular}
\end{table*}

\section{Minimum variance threshold}~\label{sec:split-threshold}

In this paper, we propose the Minimum Variance Threshold (MVT) to replace MAD as a selection criteria for the static and dynamic $\epsilon$-lexicase algorithms. The MVT split the errors into two clusters, reducing the pool of individuals to those below the specific threshold. This criterion is inspired by regression trees \cite{breiman1984classification}.
Essentially, the parent selection method changes the selection distribution to specific cases, and we hypothesize that performing the selection based on clustering good and bad performing individuals can improve the overall performance of $\epsilon$-lexicase selection.

In each step of MVT lexicase, the threshold $\tau*$ for staying in the pool is such that the sum of the variance of the errors in the pool on either side of the threshold is minimized. To estimate the threshold $\tau*$, we calculate:

\begin{equation}~\label{eq:split_threshold}
    \tau^* = \underset{\text{min}(\mathbf{e}_t) < \tau < \text{max}(\mathbf{e}_t)}{\text{min}} \left ( \frac{\text{Var}(\mathbf{l})}{|\mathbf{l}|} + \frac{\text{Var}(\mathbf{r})}{|\mathbf{r}|} \right ),
\end{equation}

\noindent where $\mathbf{l} = [e_t : \mathbf{e}_t | e_t < \tau]$ and $\mathbf{r} = [e_t : \mathbf{e}_t | e_t \geq \tau]$ are the left-sided and right-sided partition, respectively. 

\begin{figure}[tbh]
    \centering
    \includegraphics[width=\linewidth]{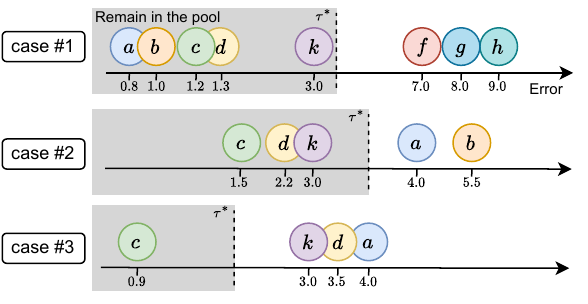}
    \caption{Process of consecutively splitting the pool of individuals into two clusters based on their error ($x$ axis). The process consists of randomly picking a test case, estimating $\tau^*$ by solving Eq. \ref{eq:split_threshold}, and removing individuals with errors higher than the pool threshold. This is repeated until only one individual remains in the pool, or all training data was already used as singular test cases --- returning one random individual from the remaining pool.}
    \label{fig:feat_splis}
\end{figure}

The procedure is to search for a value that splits the errors by clustering them into one group with smaller error values (left) and one with greater error values (right). Figure \ref{fig:feat_splis} illustrates this process. Every time we perform a split, the remaining pool represents all individuals who consistently performed better in the error cases (that is, on the left side of the split). We see this process as a more intuitive way of splitting the pool.

\section{Feature Engineering Automation Tool}~\label{sec:feat}

The Feature Engineering Automation Tool (FEAT) was proposed in \cite{la_cava_learning_2019} as an application of symbolic regression and was previously benchmarked in SRBench with good results \cite{srbench}. The idea is to evolve a set of trees used as features in another machine learning (ML) model.
We used FEAT as our SR framework to evaluate our proposed method. Our choice is motivated by semi-dynamic $\epsilon$-lexicase being the default selection method and, additionally, because the model was studied in previous works with different lexicase methods \cite{la_cava_learning_2019}, also successfully applied to clinical decision modeling \cite{la_cava_flexible_2023}.

The algorithm follows the standard steps of an evolutionary algorithm, with some specific particularities besides the $\epsilon$-lexicase selection. First, it performs a multi-objective optimization, minimizing both the fitness and complexity of models. Second, it represents individuals as a collection of symbolic regression trees. Third, it uses a backpropagation algorithm to optimize the parameters of the models. While the selection is made with $\epsilon$-lexicase, survival of the population is done using the non-dominated sorting genetic algorithm (NSGA2) \cite{NSGA-II}.

\begin{figure}[tbh]
    \centering
    \includegraphics[width=.85\linewidth]{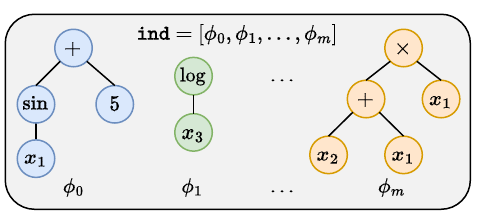}
    \caption{An individual in FEAT is a collection of symbolic regression trees as meta-features for any machine learning model.}
    \label{fig:feat_individual}
\end{figure}

Each individual is the combination of $\phi_0, \phi_1, \ldots, \phi_m$ expression trees representing one new feature. These features are used as inputs for any machine learning model (using a linear regression with $l_2$ regularization \cite{ridge} by default). Figure \ref{fig:feat_individual} illustrates a possible individual.

The evolutionary loop of FEAT performs the evolution of a population of individuals as an application of evolutionary computing for feature engineering. The fitness of an individual is calculated by taking an ML model $f$ and training it with the collection of features corresponding to that individual, then evaluating the mean squared error between the predictions of the ML model $\mathbf{\widehat{y}} = f([\phi_0, \phi_1, \ldots, \phi_m])$ and observed values $\mathbf{y}$:

\begin{equation}
    \text{MSE}(\mathbf{\widehat{y}}, \mathbf{y}) = \frac{1}{d} \sum_{i=1}^{d} \left ( \widehat{y}_i - y_i \right )^2.
\end{equation}

The complexity of a model is defined recursively for a node $n$ with $k$ arguments as the combination of the complexity of its children with the complexity of the node itself: 

\begin{equation}
    C(n) = c_n * (\sum_{a=1}^k C(a)).
\end{equation}

The complexity of an individual is the sum of the complexity for each feature $\phi$ it contains, and the model size is the number of nodes for each feature $\phi$. The complexity of each feature increases exponentially as the depth increases but tends to reach higher values if deeper nodes are more complex. This way, the occurrence of complicated operators is discouraged, and we focus on more interpretable models.

Crossover and mutation are implemented as follows. Crossover can either swap two subtrees between two individuals or swap two features (whole trees). For mutations, the algorithm uses \textit{point}, \textit{insert}, and \textit{delete} mutations (randomly replace, insert, or delete one node, respectively, as well as \textit{insert/delete} dimension, that creates or removes a new tree representing a dimension.

FEAT implements two mechanisms to split the training data further. The first, called \textit{validation split}, internally separates the training data into two partitions: one for fitting the parameters and evaluating the model and validation data for logging and picking the final individual from the last population. The second, called \textit{batch learning}, generates a random batch from the inner training partition to perform the fit and can be used to implement the downsampling strategy proposed in \cite{geiger_down-sampled_2023}.

Mainly related to lexicase, previous work showed that downsampling the test cases can achieve similar or better performance \cite{10.1162/evco_a_00346, 10.1145/3319619.3326900, 10.1145/2908812.2908851}, and $\epsilon$-lexicase was also shown to hold these benefits \cite{geiger_down-sampled_2023}. 

\section{methods}~\label{sec:methods}

Our experiments are twofold. First, we evaluate the FEAT algorithm with our proposed parent selection schema with $30$ runs for $6$ datasets. For the first set of experiments, we focus on performing longer runs and obtaining a larger sample size to calculate statistics and perform an in-depth analysis. We changed the NSGA2 survival step by replacing the original population with the offspring to isolate the effect of parenting selection, as NSGA2 is an elitist algorithm. The batch size was also turned off to assess the number of test cases each variation performed.

After the first batch of experiments, we evaluate it through a rigorous benchmark of symbolic regression algorithms known as SRBench \cite{srbench}, containing $122$ black-box regression problems, taken from the Penn Machine Learning Benchmarks (PMLB) \cite{romano2021pmlb}, which include the Friedman problems \cite{4a848dd1-54e3-3c3c-83c3-04977ded2e71} ($62$ datasets). The Friedman contains synthetic problems designed to be tricky to solve by regression models, while the rest of the datasets contain real-world data. For the SRBench, we performed $10$ runs for each dataset, as specified by the benchmark. We kept every hyper-parameter to correspond to the benchmark settings, allowing us to compare our results with the latest published result.

\begin{figure}[tbh]
    \centering
    \includegraphics[width=.75\linewidth]{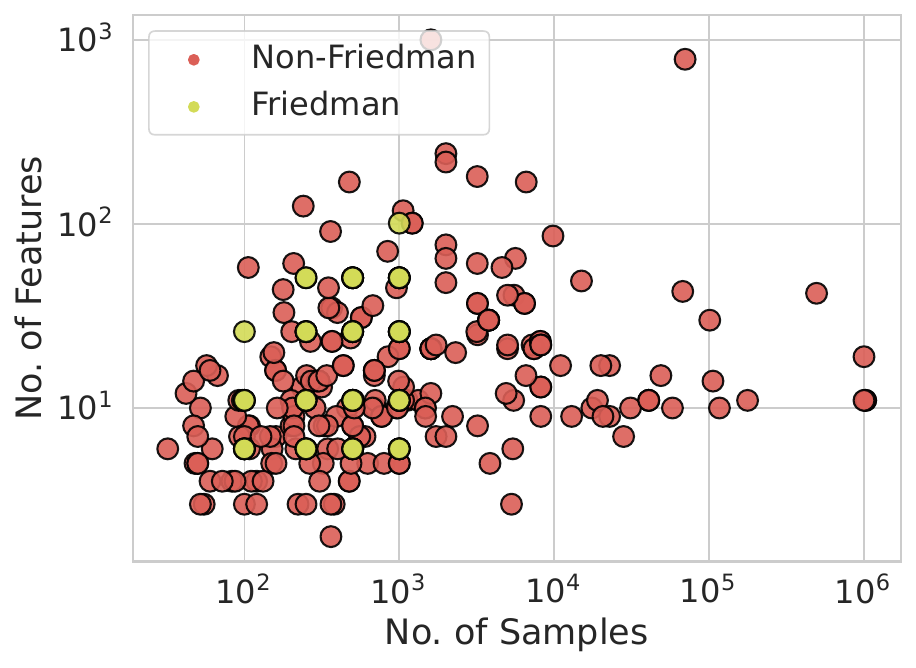}
    \caption{Dimensionality of the SRBench datasets.}
    \label{fig:pmlb_size}
\end{figure}

Regardless of whether it was from the first or second batch of experiments, every run was done by splitting the data into train and test partitions in a ratio of $.75/.25$. Table \ref{tab:datasets} and Figure \ref{fig:pmlb_size} show the dimensionality of each dataset used in the experiments. 

\begin{table}[tbh]
    \centering
    \caption{Dimensionality of the six datasets used to perform an in-depth analysis.}
    \label{tab:datasets}
    \begin{tabular}{lll}
    \hline
    \textbf{Dataset}        & \textbf{\# samples} & \textbf{\# features} \\ \hline
    airfoil        & $1503$       & $5$           \\
    concrete       & $1030$       & $8$           \\
    energy cooling & $768$        & $8$           \\
    energy heating & $768$        & $8$           \\
    Housing        & $506$        & $13$          \\
    Tower          & $3135$       & $25$          \\ \hline
    \end{tabular}
\end{table}

Table \ref{tab:feat-hyperparameters} shows the hyper-parameter configuration used in the SRBench experiment, with values in parenthesis showing the settings used in the first $6$ datasets. The batch size was selected after reading the results obtained from the first batch of experiments. The complexity for each operator specified in the hyper-parameter is depicted in Table \ref{tab:complexity}, reminding the reader that these values are pre-defined by the FEAT package authors.

\begin{table}[tbh]
    \centering
    \caption{FEAT hyper-parameters shared between all evaluated variations. Values in parenthesis indicates a configuration used just with the $6$ datasets.}
    \label{tab:feat-hyperparameters}
    \begin{center}
    \begin{tabular}{ll}
        \toprule
        \textbf{Parameter} & \textbf{Value} \\
        \midrule
        objectives & ["fitness","complexity"] \\
        pop\_size & $100$ \\
        gens & $100$ ($350$) \\
        cross\_rate & $0.5$ \\
        ml & Linear ridge regression \\
        max\_depth & $6$ ($3$) \\
        backprop & True \\
        iters & $10$ \\
        validation split & $0.25$ \\
        selection & NSGA2 (offspring) \\
        batch\_size & $200$ (unlimited) \\
        functions & [$+$,$-$,$*$,$\frac{\cdot}{\cdot}$,$(\cdot)^2$,$(\cdot)^3$,$\sqrt{\cdot}$,$\text{sin}$,\text{cos},$e^{(\cdot)}$,$\text{log}$] \\
        \bottomrule
    \end{tabular}
    \end{center}
\end{table}

\begin{table}[tbh]
    \centering
    \caption{Complexity of each operator}
    \label{tab:complexity}
    \begin{tabular}{@{}ll@{}}
    \toprule
    \textbf{Complexity} & \textbf{Operators} \\ \midrule
    1          & $+$, $-$       \\
    2          & $\frac{\cdot}{\cdot}$, $*$, $(\cdot)^2$, $\sqrt{\cdot}$      \\
    3          & $\text{cos}$, $\text{sin}$, $(\cdot)^3$     \\
    4          &  $e^{(\cdot)}$, $\text{log}$         \\ \bottomrule   
    \end{tabular}
\end{table}

Original FEAT implementation is available at \url{https://github.com/cavalab/feat}. FEAT with the MVT selection is available at \url{https://github.com/gAldeia/feat}. All data and the source code for implementations, experiments, and post-processing analysis are available at \url{https://github.com/gAldeia/srbench/tree/feat_split_benchmark}.

\section{results and  discussion}~\label{sec:results-discussion}

We divide this section into three parts.
First, we analyze the convergence aspects of the different $\epsilon$-lexicase algorithms, focusing on how they affect the search convergence and the number of test cases they use. 
Then, we benchmark the proposed method with the SRBench framework, which consists of $23$ machine learning methods (originally $21$ plus our $2$ proposed algorithms), of which $16$ ($14$ plus ours) are symbolic regression algorithms.
Finally, we also use the SRBench to asses how the proposed method scales with the number of features and number of samples of the dataset. 

As for naming, FEAT($\epsilon$-lex) stands for the standard FEAT with default $\epsilon$-lexicase selection. D-Split and S-Split are our dynamic and static methods of estimating the threshold using FEAT.

\subsection{Behavior during the run}

\begin{figure}[tbh]
    \centering
    \includegraphics[width=\linewidth]{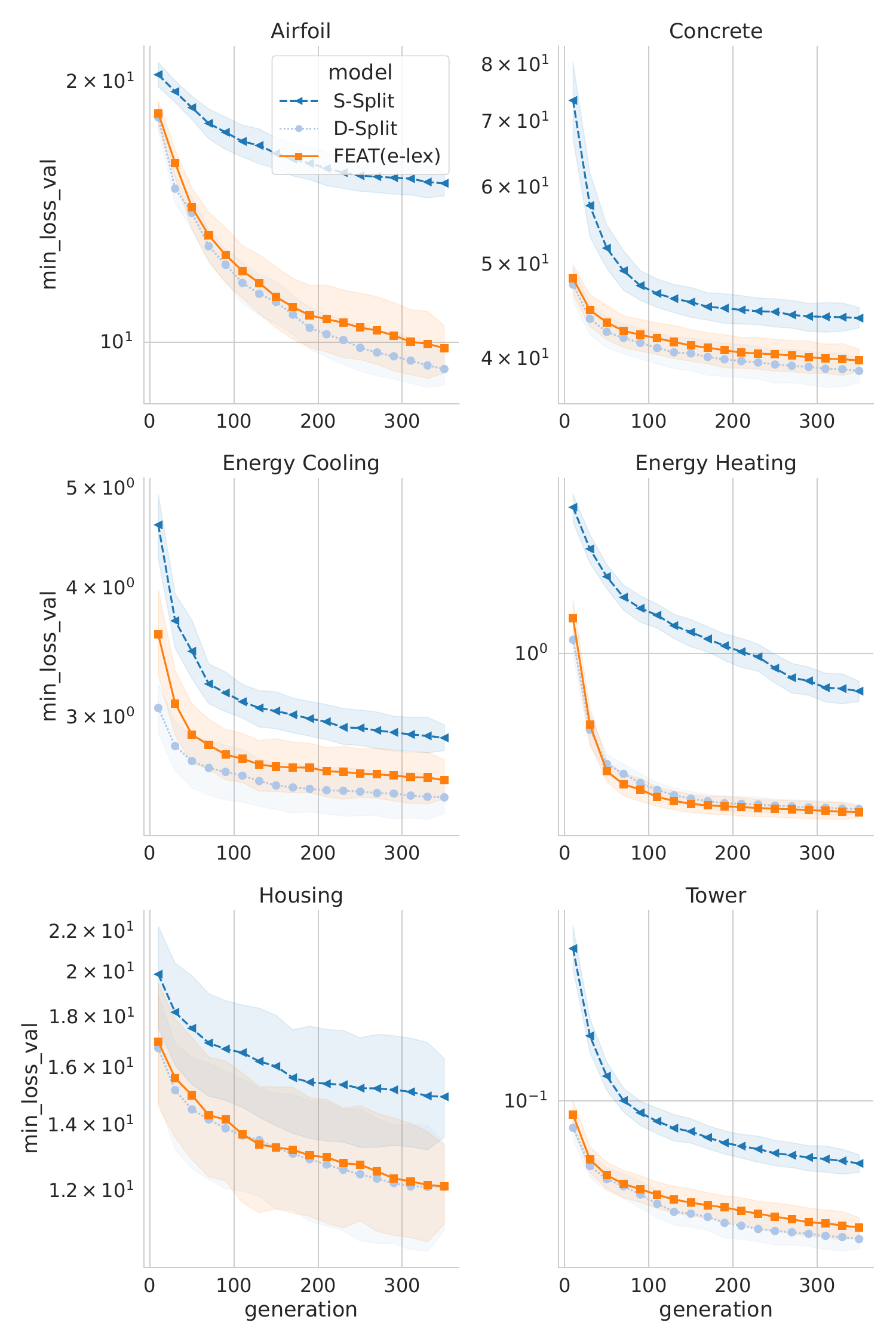}
    \caption{Convergence loss of the best individual on validation partition for the six problems.}
    \label{fig:min_loss_val}
\end{figure}

\begin{figure}[tbh]
    \centering
    \includegraphics[width=\linewidth]{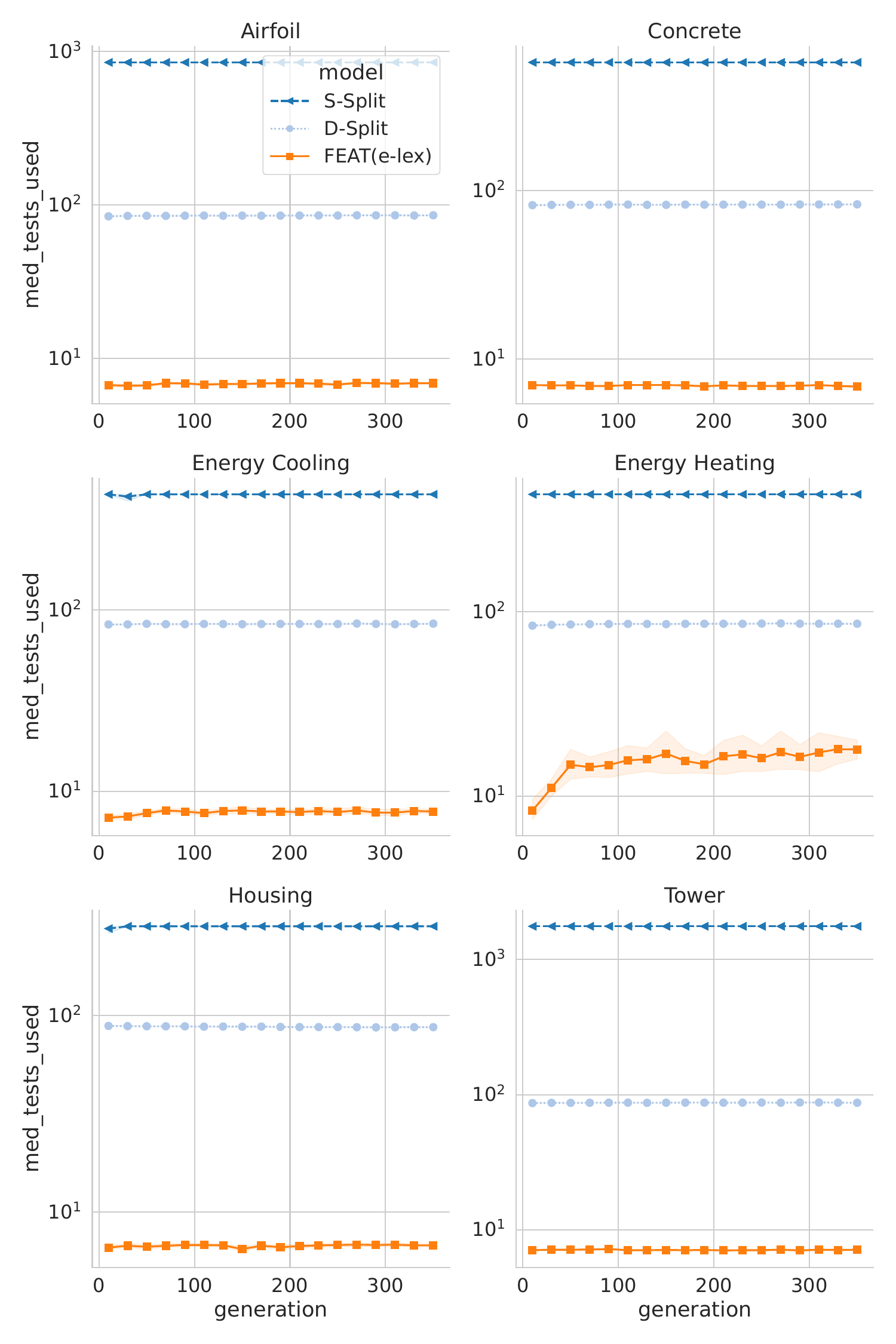}
    \caption{Median number of test cases used to pick each parent for the six problems.}
    \label{fig:max_tests_used}
\end{figure}

Figure \ref{fig:min_loss_val} reports the minimum loss on the validation split for each dataset. Figure \ref{fig:max_tests_used} reports the median number of test cases used to select the parent pool for each generation.

The D-Split shows a better convergence curve in two datasets (Airfoil and Energy Cooling) than other approaches, supporting the idea that the MVT can change the convergence curves.
S-Split shows the worst convergence curve for all datasets. In some cases, such as in Concrete, Energy Cooling, and Tower datasets, it seems to reach a \textit{plateau} with a higher error rate.

When looking at the number of test cases used in each generation, we notice that for all the datasets, the original $\epsilon$-lexicase requires a smaller number of cases, implying a faster execution time (as we will discuss later). 
Both D-Split and S-Split used more test cases, with S-Split using twice as many test cases as D-Split. This implies that many test cases have large clusters of good-performing individuals that would be ignored in the selection using the MAD, but with the MVT, they are kept in the pool for longer. We believe that this creates a more robust selection; thus, we have improvements over the original FEAT with the MAD threshold.

We believe that clustering the whole population can lead to high thresholds and a less effective selection pool, and the S-Split could be improved by implementing a down-sampling strategy that selects a subset of test cases where it is more likely to have significant differences between the individuals \cite{10.1162/evco_a_00346}.
The D-split ---which uses the pool to calculate the threshold--- yielded a better performance because the pool shrinks, removing individuals with a subpar overall goodness-of-fit and getting more stable results.

\subsection{Performance on small datasets}

\begin{figure}[tbh]
    \centering
    \subfloat[Median $R^2$ rank\label{barplot_r2_rank}]{%
       \includegraphics[width=0.5\linewidth]{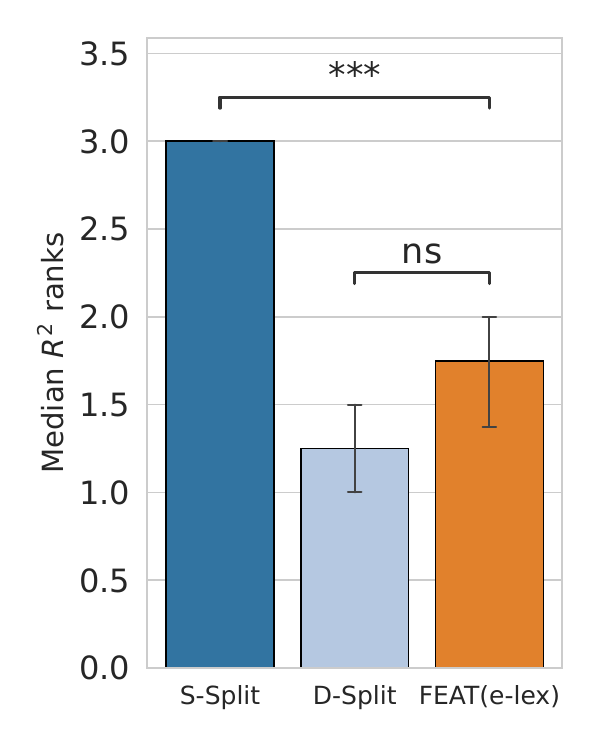}}
    \subfloat[$R^2$ score\label{boxplot_r2_score}]{%
       \includegraphics[width=0.5\linewidth]{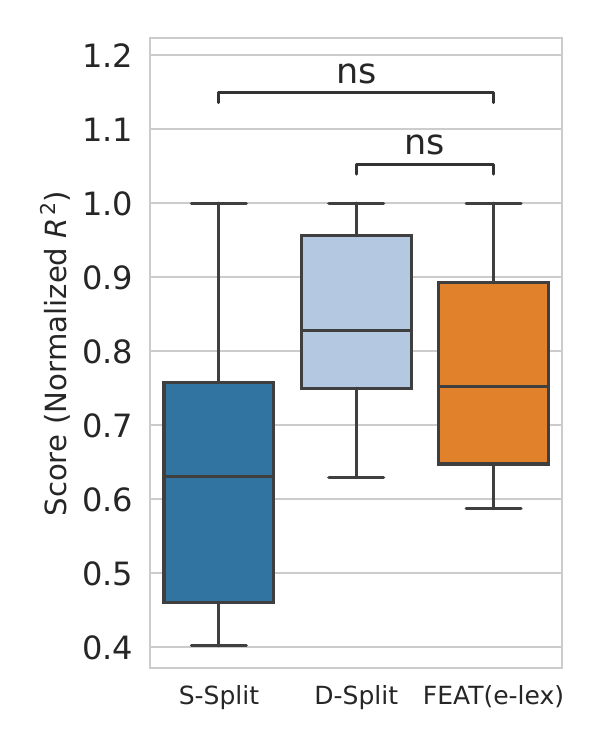}}
    \\
    \subfloat[$R^2$ critical difference diagram\label{r2_cd}]{%
       \includegraphics[width=\linewidth]{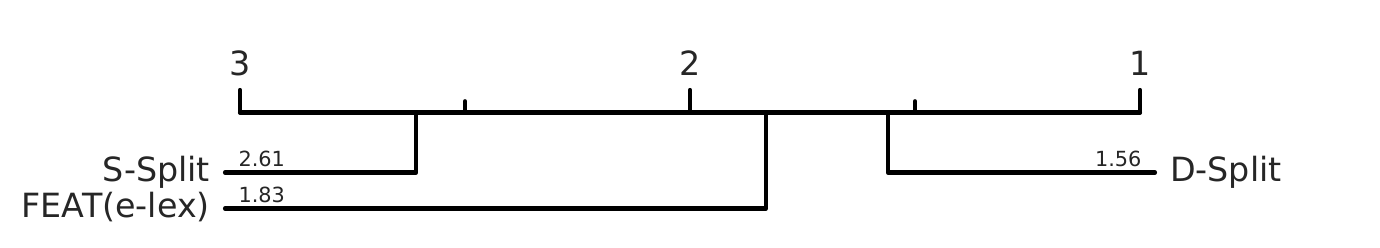}}
    \caption{Grouped $R^2$ results for the $6$ datasets.}
    \label{fig:r2_results_group}
\end{figure}

\begin{figure}[tbh]
    \centering
    \subfloat[Sizes\label{boxplot_size}]{%
       \includegraphics[width=0.5\linewidth]{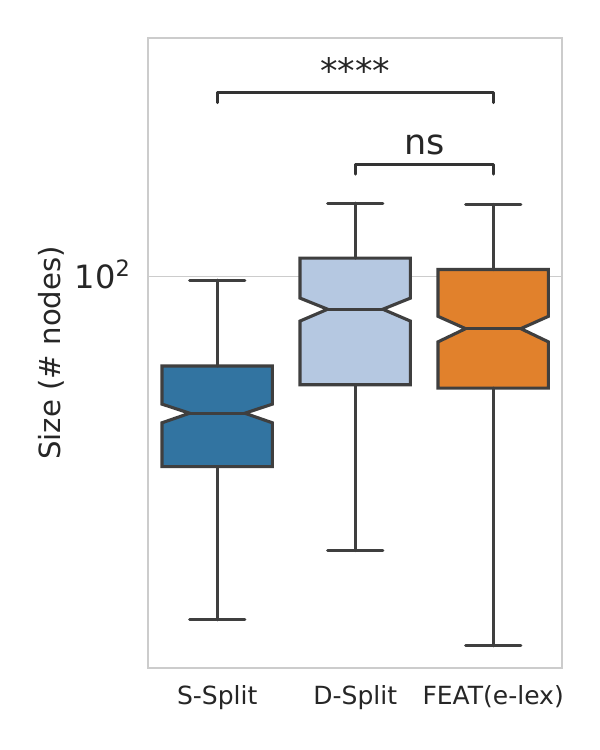}}
    \subfloat[Complexites\label{boxplot_complexity}]{%
       \includegraphics[width=0.5\linewidth]{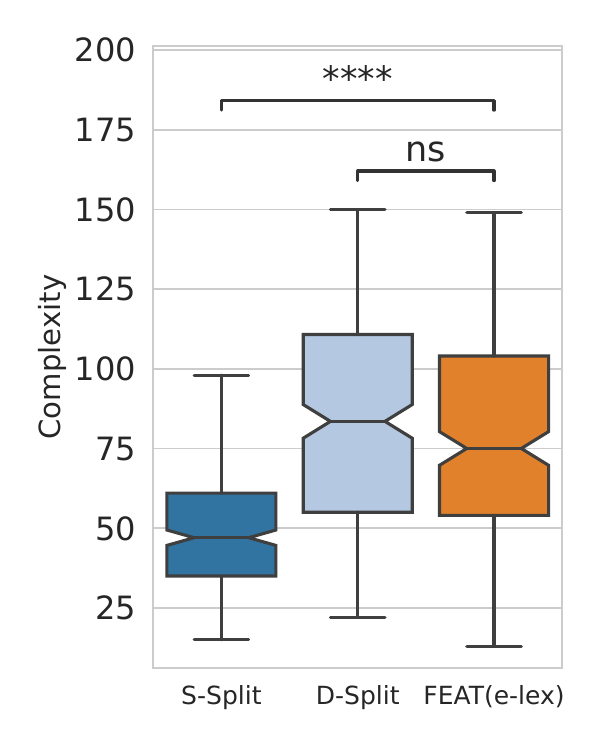}}
    \\
    \subfloat[Complexity critical difference diagram\label{complexty_cd}]{%
       \includegraphics[width=\linewidth]{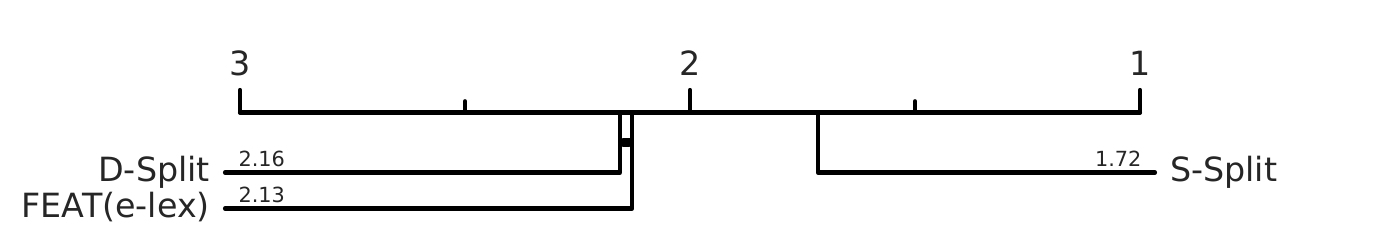}}
    \caption{Grouped complexity results for the $6$ datasets.}
    \label{fig:complexity_results_group}
\end{figure}

To assess the final performance for the first $6$ datasets, we report in Figure \ref{fig:r2_results_group} the median $R^2$ ranking for each random seed used in the experiments, and in Figure \ref{fig:complexity_results_group}, we show the size and complexity of the final models. We also included a critical differences diagram \cite{IsmailFawaz2018deep}, showing each algorithm's median ranking, with horizontal lines connecting them when no statistical significance is observed from the samples.

Looking at the $R^2$ rank (Figure \ref{barplot_r2_rank}) we see S-Split as the worse variant, with $3$ asterisks indicating $\text{p-value} \leq 1\times 10^{-3}$. The critical differences (\ref{r2_cd}) shows that D-Split is mainly classified as first among the other variants. When looking at the grouped $R^2$ for all datasets (Figure \ref{boxplot_r2_score}), there is no statistical difference, but $75\%$ of the distribution of the D-Split is above the median for the original $\epsilon$-lexicase.
The worst performance is from the S-Split, with a large spread between $[0.3, 0.8]$.

The worst performance of S-Split shows a better trade-off in terms of size and complexity (Figures \ref{boxplot_size} and \ref{boxplot_complexity}).
The critical differences diagram shows that, while S-Split is the best algorithm in size and complexity, the other two variants, D-Split and $\epsilon$-lexicase, have no statistical difference.

\subsection{Benchmarking with SRBench}

\begin{figure*}[tbh]
    \centering
    \includegraphics[width=.9\linewidth]{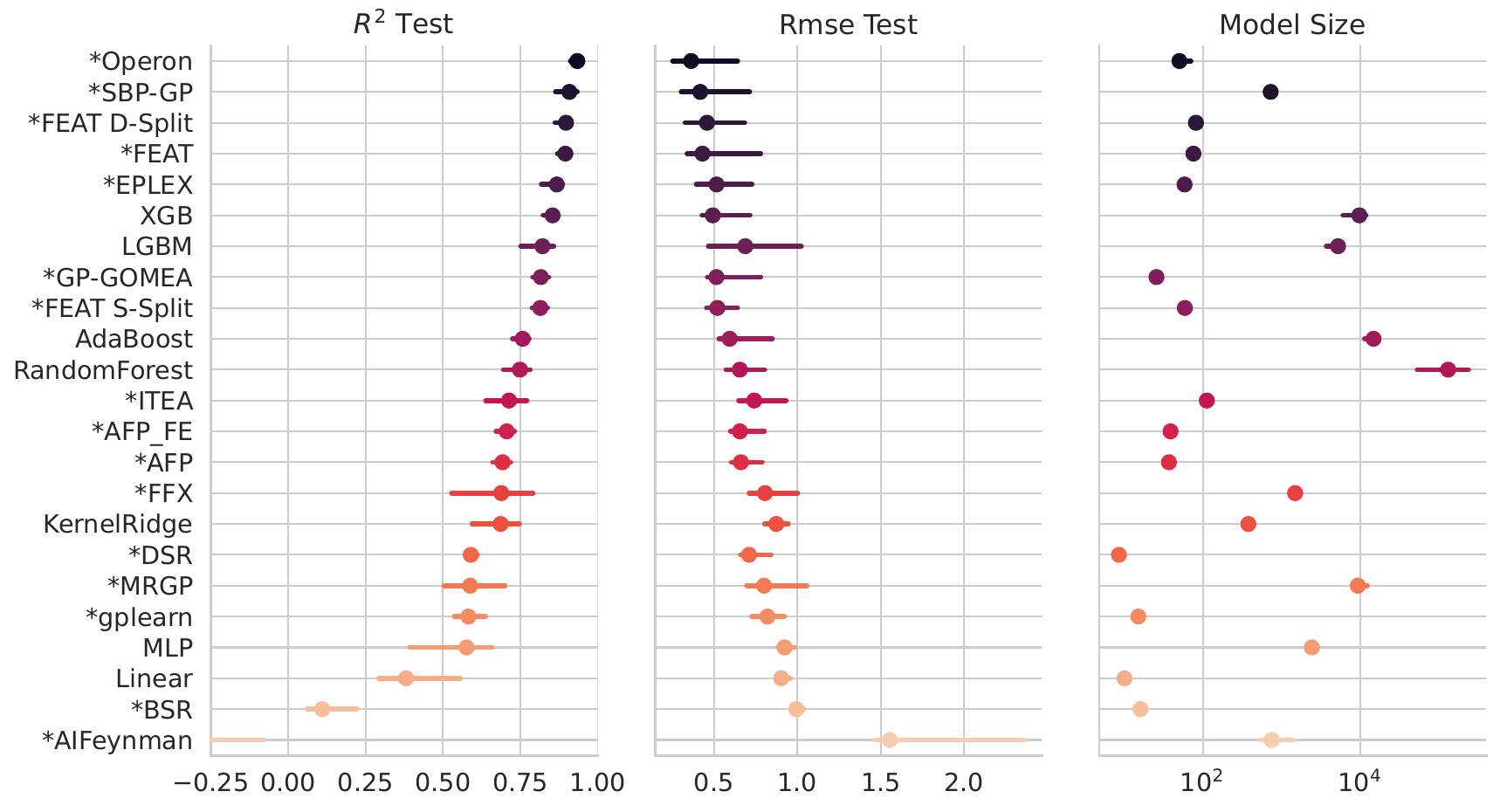}
    \caption{Comparison of our proposed algorithms with SRBench results. Names with an asterisk indicate that the model is a symbolic regression algorithm. Bars show the $95\%$ confidence interval. Size comparisons are made by the number of nodes to keep models comparable.}
    \label{fig:srbench}
\end{figure*}

The SRBench framework comprises several machine learning models, some of which are symbolic regression algorithms, meaning they could benefit from our proposed lexicase selection. We notice that, from the last published analysis of state-of-the-art symbolic regression methods, FEAT was one of the best-performing algorithms, originally ranked at $3$ in terms of the root of $R^2$ on test partition, behind Operon \cite{Kommenda2019} and SBP-GP \cite{10.1145/3321707.3321758}. Figure \ref{fig:srbench} reports the latest available results from SRBench, combined with our results of running our two variations, FEAT S-Split and FEAT D-Split, over the benchmark problems.

The FEAT D-Split outperforms the original FEAT by a modest difference, showing approximately similar RMSE, $R^2$, and model size.
FEAT S-Split performs worse than the other two variations of FEAT but still ranks 9th, outperforming more than half of the methods.

\begin{figure}[tbh]
    \centering
    \includegraphics[width=\linewidth]{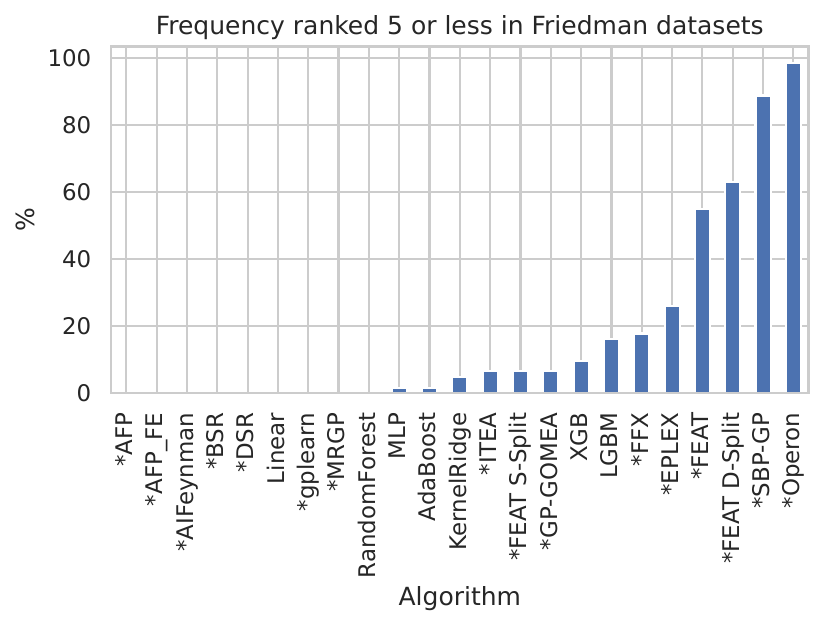}
    \caption{Number of times each algorithm was ranked top 5 or less for each run for the Friedman datasets.}
    \label{fig:hist5fri}
\end{figure}

\begin{figure}[tbh]
    \centering
    \includegraphics[width=\linewidth]{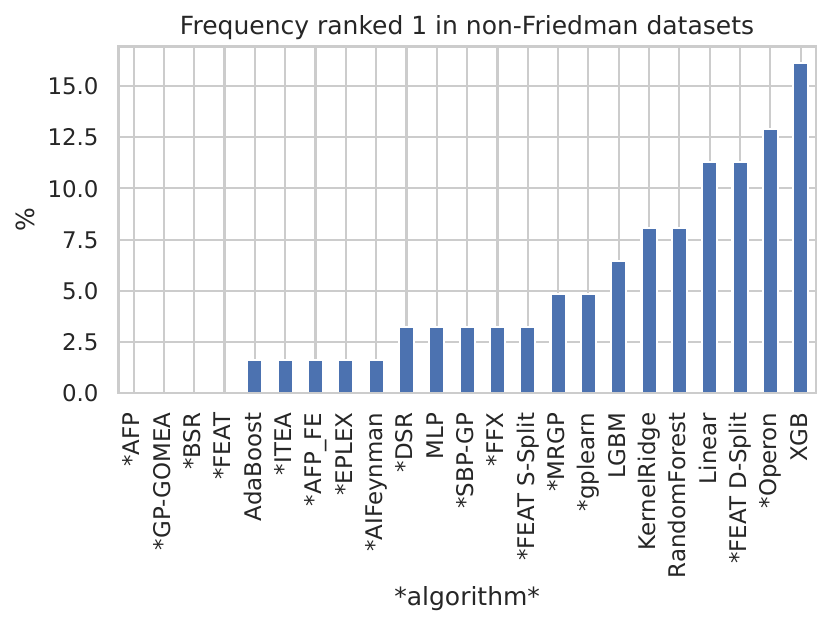}
    \caption{Number of times each algorithm was ranked top 1 or less for each run for the non-Friedman datasets.}
    \label{fig:hist1nonfri}
\end{figure}

As pointed out in \cite{de2023transformation}, the aggregated result of SRBench can mask the difference among the top algorithms. A division of the benchmarks into Friedman and non-Friedman datasets (roughly half and half of the datasets) revealed that the best-ranked algorithms had a noticeable difference on the Friedman benchmarks. At the same time, they all behaved similarly in the non-Friedman sets. Following \cite{de2023transformation}, we report in Figure \ref{fig:hist5fri} the percentage of times that each of the algorithms was ranked top 5 for the Friedman synthetic problems, and Figure \ref{fig:hist1nonfri} reports the percentage each algorithm was ranked top 1 in the real-world, non-Friedman problems. This represents a relative goodness-of-fit performance when pairing each instance of results, and higher values are better.
We did not report the top 1 percentage for Friedman datasets because no FEAT variation reached more than $4\%$ of wins, implying that these problems are still challenging for FEAT.

While FEAT achieves a rank of 5 or less approximately $53\%$ of the time, our proposed method, FEAT D-Split, can increase this percentage to over $60\%$.
FEAT with S-Split selection is among the top 5 algorithms in less than $10\%$ of the runs.
The D-Split performs better than the other variants, with a total of $10\%$ of wins among the other algorithms. This represents a significant improvement, as the original FEAT performs poorly in this set of problems.
We also noticed that S-Split performed better for the non-Friedman datasets than the original FEAT. Given that S-Split was better at finding models with smaller size and complexity, this implies that it found solutions with a better trade-off between $R^2$ and complexity.

Overall, these histograms show that FEAT D-Split increased its rank in the Friedman benchmarks by a small margin, but as already noticed, not enough to be among the top-$1$ ranks. 
D-Split and S-Split significantly increase solutions at the top-$1$, with D-Split ranking in the top-$3$ algorithms for the non-Friedman benchmarks. Again, as noted in \cite{de2023transformation}, the non-Friedman median $R^2$ does not have much influence on the overall rank. Thus, we could not observe much difference in Fig.\ref{fig:srbench}.
Notice that this plot discretizes the results presented in the previous plots. The fact that it was not ranked first in a particular instance does not imply that the returned result is not close to the best result. This is better highlighted on the error bar plots.

\begin{figure}[tbh]
    \centering
    \includegraphics[width=\linewidth]{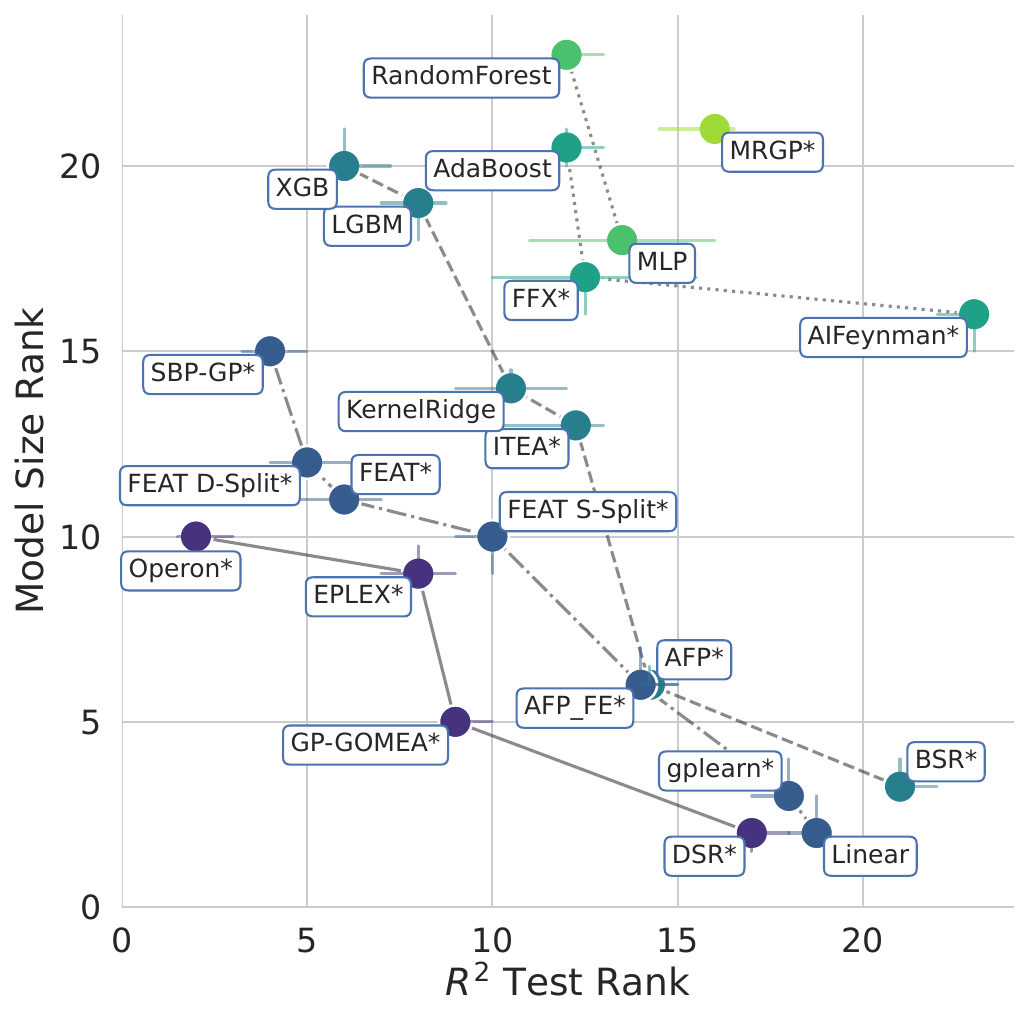}
    \caption{Pareto front of benchmarked algorithms}
    \label{fig:pareto_plot_srbench}
\end{figure}

Finally, we plot the rank for model size and $R^2$ for all algorithms in SRBench in Figure \ref{fig:pareto_plot_srbench}. For both model size and $R^2$, smaller values are better. The lines in the plot represent the Pareto fronts for each rank. 

We can see that all variations are within the same Pareto front rank, with different levels of trade-off between model size and accuracy.
While showing equivalent performance in the synthetic benchmarks, we saw that the lexicase selection achieved a better result in real-world problems.

\subsection{Scalability}

\begin{figure}[tbh]
    \centering
    \includegraphics[width=.75\linewidth]{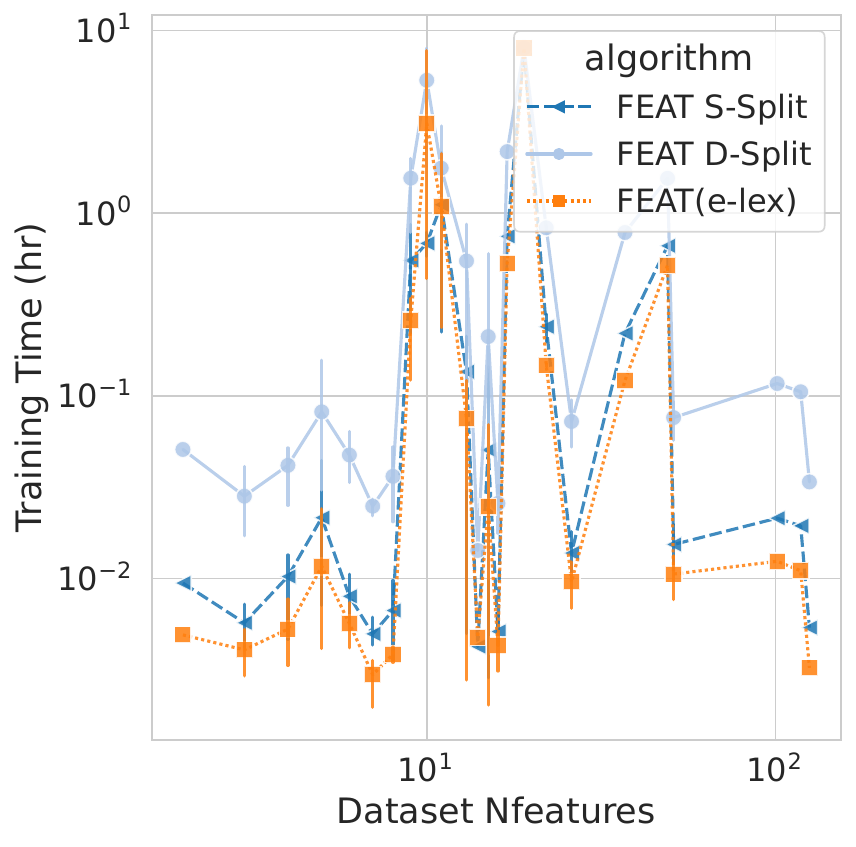}
    \caption{Training time versus number of features in the dataset}
    \label{fig:training_time_dataset_nfeatures}
\end{figure}

\begin{figure}[tbh]
    \centering
    \includegraphics[width=.75\linewidth]{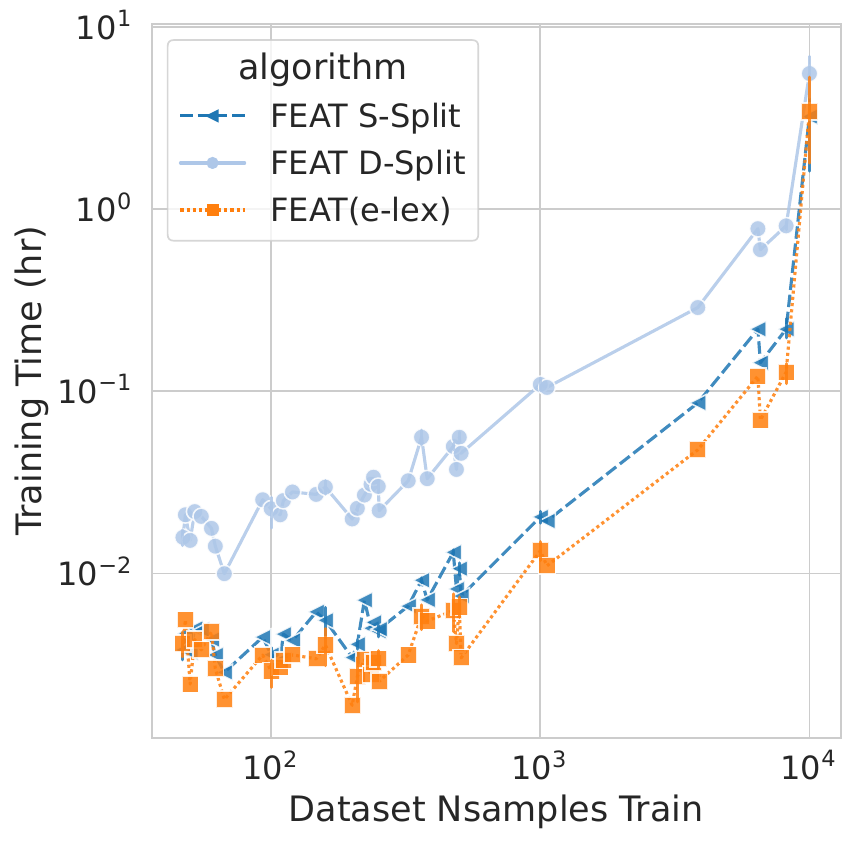}
    \caption{Training time versus number of samples in the training partition of the dataset}
    \label{fig:training_time_dataset_nsamples}
\end{figure}

By plotting the execution time versus number of samples or features for the datasets in SRBench, we can visualize how different implementations scale. We re-run the entire FEAT algorithm with SRBench datasets to remove hardware-related differences. Figure \ref{fig:training_time_dataset_nfeatures} reports the training time for different numbers of features in the dataset, and Figure \ref{fig:training_time_dataset_nsamples} reports the training time for different numbers of samples.
We did not compare execution time for other models as they are hardware-dependent. Thus, we could get a biased estimate of performance.

All three variants of FEAT with different parent selections have similar curves but different offsets.
The results imply that scalability is not correlated with the number of features but with the number of samples.
This may be explained because it increases the number of test cases, and our MVT selection uses a higher number of test cases than the original $ \epsilon $ -lexicase.

As for the number of features, the size of the individuals is limited by the maximum size of symbolic regression trees, and, as with many symbolic regression algorithms, they have to perform the feature selection implicitly due to size constraints. While the number of features does not impact training time, the maximum expression sizes do. However, we achieved a good performance on SRBench with relatively small models, showing no need to increase the maximum allowed size of the models and implying FEAT could perform well training with the current settings.

One interesting result is that S-Split uses more tests than D-Split but still has a faster execution. We hypothesize that this is because D-Split needs to calculate the threshold for each test case until one parent is selected, while S-Split needs to calculate it only once.

\section{Conclusions}~\label{sec:conclusions}

In this paper, we proposed a novel way of estimating the threshold to stay in the pool for $\epsilon$-lexicase selection, an adaptation to improve lexicase for regression problems.
We then evaluated these new threshold criteria with $6$ small problems and $122$ regression problems, comparing the results with other $21$ algorithms.
We achieve competitive results with the previous version of $\epsilon$-lexicase while showing improvements in real-world datasets. The performance in synthetic problems was preserved.
The downside of our implementation is that it requires more test cases to run, leading to higher execution times. Nevertheless, using additional resources helps achieve better results, as shown by the final performance comparisons.

In future work, we plan to study how these new criteria change the distribution probability of individuals in the population. We will also investigate how down-sampling can decrease the number of test cases used. Finally, these criteria can also be implemented into other symbolic regression algorithms, and we can assess whether they can be improved with the minimum variance split criteria.

\begin{acks}
W.G.L. was supported by National Institutes of Health (NIH) grant R00-LM012926, and Patient Centered Outcomes Research Institute (PCORI) ME-2020C1D-19393.
F.O.F. is supported by Funda\c{c}\~{a}o de Amparo \`{a} Pesquisa do Estado de S\~{a}o Paulo (FAPESP) grant 2021/12706-1, and Conselho Nacional de Desenvolvimento Cient\'{i}fico e Tecnol\'{o}gico (CNPq) grant 301596/2022-0.
G.S.I.A. is supported by Coordena\c{c}\~{a}o de Aperfei\c{c}oamento de Pessoal de N\'{i}vel Superior (CAPES) finance Code 001 and grant 88887.802848/2023-00.
\end{acks}

\bibliographystyle{ACM-Reference-Format}
\bibliography{sample-base}

\end{document}